\documentclass[letterpaper,10pt,conference]{ieeeconf}

\usepackage{graphicx}
\usepackage{booktabs}
\usepackage{amsmath}
\usepackage{amssymb}
\usepackage[table]{xcolor}

\title{\LARGE \bf
Mind-VLA: Instruction-Aware Spatial Representation Alignment for
Vision-Language-Action Models
}

\author{
  Xingyu Ding\textsuperscript{1,2} \quad
  Yuzhong Zhao\textsuperscript{3} \quad
  Yang Wu\textsuperscript{2,3} \quad
  Chunhai Zhao\textsuperscript{2}  \\
  Chaoyang Zhao\textsuperscript{2,$\dagger$} \quad
  Yifan Zhang\textsuperscript{2,3,$\dagger$} \quad
  Jian Cheng\textsuperscript{2,3} \\
  \textsuperscript{1}Nanjing University, Nanjing, China \\
  \textsuperscript{2}Institute of Automation, Chinese Academy of Sciences, Beijing, China \\
  \textsuperscript{3}University of Chinese Academy of Sciences, Beijing, China \\
  \textsuperscript{$\dagger$}Corresponding authors
}

\begin{document}

\maketitle
\thispagestyle{empty}
\pagestyle{empty}

\begin{abstract}
Recent Vision-Language-Action (VLA) methods improve generalization by aligning their representations with 3D scene geometry.
However, these methods are fundamentally instruction-agnostic: the representations align the entire scene uniformly, neglecting the 3D geometry of the specific target
object designated by the language instruction.
This causes failures on fine-grained manipulation and target occlusion tasks, where success depends on
accurate 3D understanding of the target object rather than the entire scene. To address this, we present \textbf{Mind-VLA}, an
instruction-aware spatial representation alignment method for VLA models.
Specifically, Mind-VLA first obtains the target object specified by the language instruction, then prepares its canonical target views and extracts the corresponding VAE and VGGT features. Finally, the latent representation of the VLA model is aligned with these features to enable instruction-aware 3D understanding. 
Mind-VLA reaches 94.4\% on LIBERO and 4.47 on CALVIN with a compact 345M-parameter backbone.
On real-robot tasks with target occlusion, Mind-VLA reaches 54\% average success, outperforming the matched scene-VGGT control by 26 percentage points. 
\end{abstract}

\section{Introduction}

\begin{figure*}[tb]
    \vspace*{6pt}
    \centering
    \includegraphics[width=\textwidth]{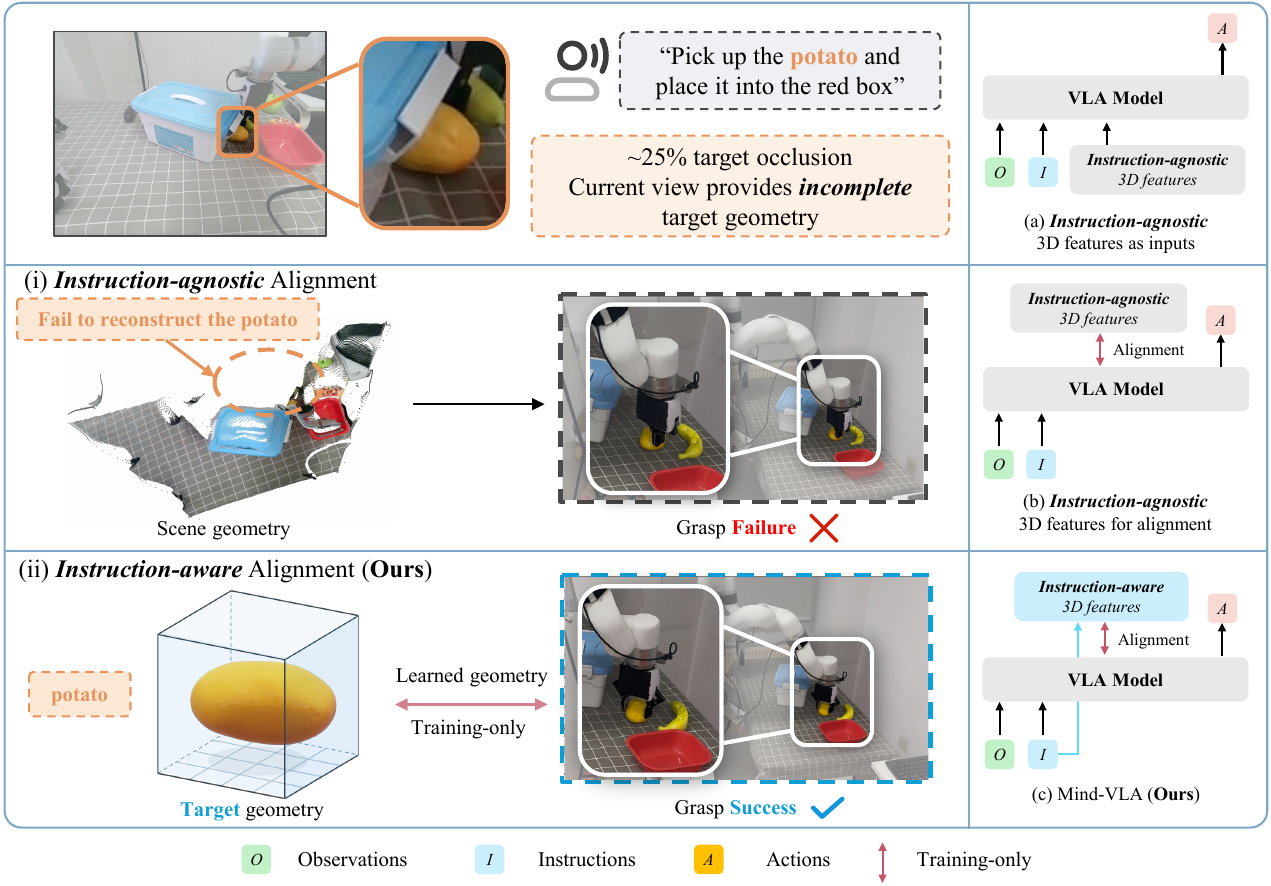}
    \caption{\textbf{Comparison of 3D-aware VLA paradigms.} The left example shows that scene-level 3D information captures incomplete target geometry under occlusion. Mind-VLA instead uses canonical target geometry for instruction-aware spatial supervision. Existing 3D-aware VLA methods are fundamentally instruction-agnostic. They either provide
    instruction-agnostic 3D features as additional model inputs (a)~\cite{geovla, pointvla, ogvla}
    or align VLA representations with instruction-agnostic 3D features during training (b)~\cite{spatialforcing, glad}, neglecting the 3D geometry of the specific target object designated by the language instruction.
    (c) Mind-VLA (Ours) aligns VLA representations with
    instruction-aware 3D features during training, enabling instruction-aware 3D understanding.}
    \label{fig:intro}
\end{figure*}

Vision-Language-Action (VLA) models have become an important approach for generalist robotic manipulation, mapping language instructions and visual observations directly to robot actions~\cite{rt2,openvla,octo,pi0}. However, many manipulation tasks require the VLA models to go beyond 2D visual recognition and reason about object shape, pose, spatial relations, and occlusion. To improve the 3D understanding of VLA models, recent methods use 3D scene information in two main ways. The first (Fig.~\ref{fig:intro}(a))~\cite{pointvla,geovla,ogvla} provides explicit 3D information as additional model input, such as point-cloud features, RGB-D observations, or rendered orthographic views. These methods can improve spatial reasoning, but they usually require depth sensors, depth estimation, or extra 3D processing during inference, increasing the deployment cost.  The second (Fig.~\ref{fig:intro}(b))~\cite{spatialforcing,glad} extracts 3D scene features with frozen 3D foundation models and aligns VLA representations to these features during training. This training-time alignment keeps the model input unchanged at inference, reduces deployment cost, and has therefore become a mainstream paradigm for building 3D-aware VLAs.

However, existing 3D-aware VLA methods are still instruction-agnostic. Their 3D input or supervision is usually built from the whole scene, regardless of which object is specified by the language instruction. As a result, the target object, distractor objects, and background structures are encoded together. This weakens the connection between the instruction and the 3D representation. For manipulation tasks, the most important geometry is often not the whole scene, but the object that the robot is asked to manipulate. When the target object is visually similar to nearby objects, instruction-agnostic supervision may make target discrimination harder. When the target is partially occluded, it may also make the VLA model less reliable, because the model has not been explicitly trained to preserve the 3D structure of the instructed object. As illustrated on the left of Fig.~\ref{fig:intro}, scene-level 3D information captures only partial target geometry under occlusion. This motivates using canonical target geometry for spatial supervision.

To address this problem, we propose Mind-VLA, an instruction-aware spatial representation alignment method for VLA models (Fig.~\ref{fig:intro}(c)). Mind-VLA aligns the VLA representation with 3D features of the instruction-specified target object instead of only using scene-level 3D features. Specifically, Mind-VLA first obtains the target object referred to by the instruction. It then prepares the canonical target views and extracts the corresponding VAE and VGGT features. Finally, the latent representation of the VLA model is aligned with these features to enable instruction-aware 3D understanding. In practice, VAE latent prediction encourages the model to encode the target object's geometry, while VGGT feature alignment guides intermediate VLA features toward target-object 3D structure. These auxiliary supervision modules are only used during training and are removed at inference time, so Mind-VLA adds negligible deployment overhead.

Our main contributions are as follows:
\begin{itemize}
    \item We identify instruction-agnostic 3D modeling as a limitation of existing 3D-aware VLA methods. Since their 3D information is built from the whole scene, they do not explicitly emphasize the object specified by the language instruction, which can hurt fine-grained manipulation and target-occlusion tasks.
    \item We propose Mind-VLA, which aligns VLA representations with target-object 3D features. Mind-VLA uses canonical target-view VAE latent prediction and target-object VGGT feature alignment during training, enabling instruction-aware 3D understanding.
    \item Mind-VLA achieves 94.4\% average success on LIBERO and 4.47 average completed length on CALVIN with a compact 345M-parameter backbone. On real-robot tasks with target occlusion, Mind-VLA reaches 54\% average success, outperforming the matched scene-VGGT control by 26 percentage points in our real-robot comparison.
\end{itemize}

\section{Related Work}

End-to-end visuomotor learning has moved from task-specific methods~\cite{diffusionpolicy, act} to generalist VLA models~\cite{rt1, rt2, octo, openvla, pi0}. Recent variants add spatial encodings~\cite{spatialvla}, visual traces~\cite{tracevla}, diffusion decoding~\cite{cogact}, and hybrid generation~\cite{hybridvla}. These methods strengthen modeling from RGB and language, but offer little explicit pressure to encode the 3D geometry of the instruction-specified object. To supply this geometry, recent work follows two directions.

The first provides explicit 3D inputs. Before VLAs, 3D manipulation methods showed the value of explicit 3D representations through voxel~\cite{peract}, point~\cite{act3d}, virtual-view~\cite{rvt}, and diffusion~\cite{3ddiffuseractor} designs, typically with multi-view RGB-D sensing and narrower task scope. Recent 3D-aware VLAs carry this to inference by feeding point clouds~\cite{pointvla, geovla}, stereo observations~\cite{stereovla}, or RGB-D orthographic views~\cite{ogvla}; while effective, they add depth or 3D preprocessing and build full-scene rather than target-object-specific geometry. The second direction instead supplies 3D information only as training supervision: Spatial Forcing (SF)~\cite{spatialforcing} aligns VLA features to VGGT~\cite{vggt} scene features, GLaD~\cite{glad} distills them across layers, and QDepth-VLA~\cite{qdepthvla} predicts quantized depth tokens. This avoids deployment-time 3D modules, but the supervision remains global: the same scene representation supervises the model regardless of the instructed object.

A related family uses predictive auxiliary objectives to enrich model representations: DreamVLA~\cite{dreamvla} predicts dynamic regions, depth, and semantic cues, PALM~\cite{palm} uses progress-aware affordances, and Geometry Forcing~\cite{geometryforcing} transfers VGGT features through video generation. Mind-VLA is complementary: it keeps inference inputs unchanged while making the supervision instruction-aware, aligning VLA representations with the canonical target views and their VGGT features rather than scene-level geometry alone.

\section{METHOD}

\begin{figure*}[tb]
\vspace*{6pt}
\centering
\includegraphics[width=\linewidth]{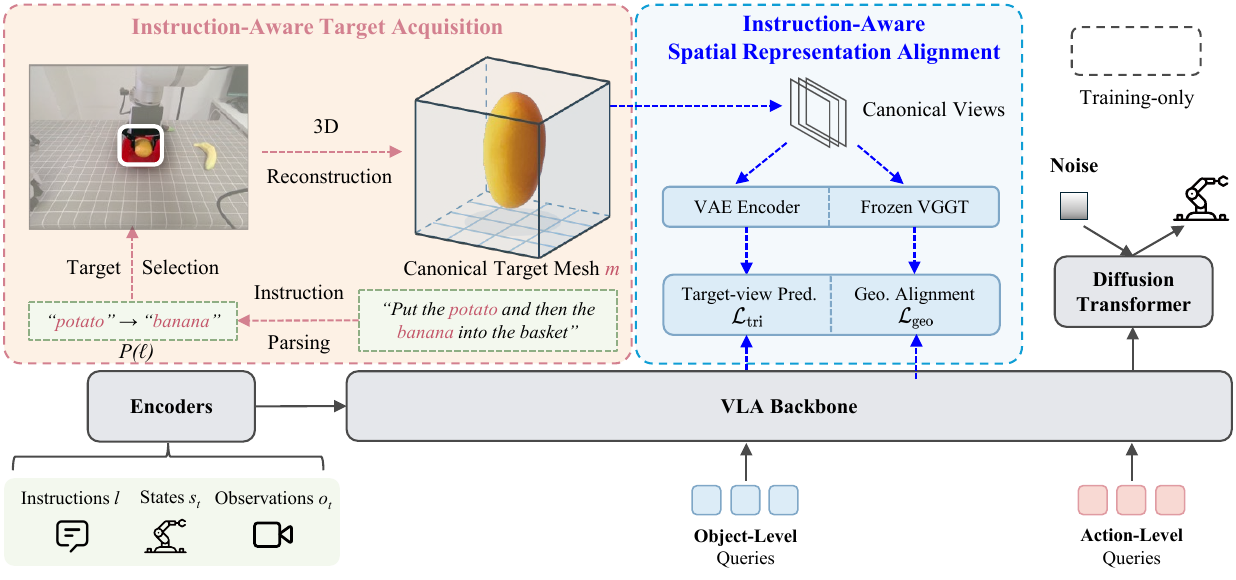}
\caption{\textbf{Mind-VLA overview.} Instructions, states, and observations are
processed by the VLA backbone for action prediction. Mind-VLA parses the
instruction into an ordered target sequence, identifies the corresponding
targets from robot demonstrations, and constructs their canonical target views
as training-only supervision. Object-level queries predict the corresponding
canonical target-view latent, while intermediate VLA representations are
aligned with target-object VGGT features.} 
\label{fig:architecture}
\end{figure*}

The overall framework of Mind-VLA is illustrated in Fig.~\ref{fig:architecture}.
We present the method in four parts. Sec.~\ref{sec:preliminaries} reviews
the basic VLA formulation and diffusion-based action generation.
Sec.~\ref{sec:target_acquisition} presents \textbf{Instruction-Aware Target
Acquisition}, which identifies instruction-specified target objects and
their manipulation order. Sec.~\ref{sec:representation_alignment} introduces
\textbf{Instruction-Aware Spatial Representation Alignment}, which constructs
canonical target views for target representation prediction and geometric
feature alignment. 
Finally, Sec.~\ref{sec:overall} describes the overall
training objective and inference procedure.

\subsection{Preliminaries}
\label{sec:preliminaries}

Vision-Language-Action (VLA) models aim to predict executable robot actions
conditioned on visual observations and language instructions, optionally
together with the robot proprioceptive state. Specifically, at
timestep $t$, the model takes the visual observation $\mathbf{o}_t$, language
instruction $\ell$, and proprioceptive state $\mathbf{s}_t$ as inputs and
jointly encodes these multimodal signals into an action-related latent
representation $\mathbf{c}_t$. Based on this representation, the action head
predicts a continuous action sequence $\mathbf{a}_{t:t+n-1}$ over the next $n$
timesteps.

For diffusion-based action generation, the action head is conditioned on
$\mathbf{c}_t$ and iteratively denoises a noisy action sequence. The action
prediction objective is defined as
\begin{equation}
\begin{aligned}
\mathcal{L}_{\mathrm{act}}
= \mathbb{E}_{\tau,\epsilon}\Big[
\big\|\epsilon - \epsilon_\theta(
&\sqrt{\bar{\alpha}_{\tau}}\mathbf{a}_{t:t+n-1} \\
&+\sqrt{1-\bar{\alpha}_{\tau}}\epsilon,
\,\tau,\,\mathbf{c}_t)
\big\|_2^2
\Big] ,
\end{aligned}
\end{equation}
where $\boldsymbol{\epsilon}_{\theta}$ denotes the action denoising network,
$\boldsymbol{\epsilon}\sim\mathcal{N}(\mathbf 0,\mathbf I)$ is Gaussian noise
added to the action sequence, $\tau$ denotes the diffusion timestep, and
$\bar{\alpha}_{\tau}$ is the cumulative coefficient of the corresponding noise
schedule. However, the action prediction objective alone does not explicitly constrain the VLA representations to capture the 3D structure of the instruction-specified target object.

\subsection{Instruction-Aware Target Acquisition}
\label{sec:target_acquisition}

To {construct} an instruction-aware VLA method, we need to identify the
target objects that are directly relevant to robot manipulation under the
language instruction. Given a language instruction $\ell$ and its corresponding
robot manipulation sequence, we define an \emph{Instruction-Aware Target} as
an object described by the instruction that needs to interact with the robot
manipulator. However, a manipulation scene may contain multiple visually or semantically
similar objects, while long-horizon tasks may involve multiple targets that
need to be manipulated in a particular order. We therefore need to identify
the actual manipulated instances and determine their manipulation order.
Specifically, given a language instruction $\ell$, e.g., ``put the potato and
then the banana into the basket,'' we first use Qwen3-VL~\cite{qwen3} to parse
the instruction and extract an ordered target phrase sequence
$\mathcal P(\ell)=\{p_m\}_{m=1}^{M}$ according to the task execution order,
e.g., $\mathcal P(\ell)=[\text{``potato''},\text{``banana''}]$. For each
target phrase $p_m$, we use a relatively low segmentation threshold and
combine the target phrase with its generic category prompt to generate a
candidate object set $\mathcal C_m$ from the robot video sequence using SAM 3~\cite{sam}.

We then identify the manipulated target using a fixed candidate-ranking strategy that combines CLIP-based semantic similarity~\cite{clip}, SAM 3 segmentation confidence, and temporal motion cues from the demonstration. For each target phrase, the highest-ranked candidate is reconstructed into a 3D mesh using SAM 3D Objects~\cite{sam3d} and used to construct the canonical target views in Sec.~III-C. The entire process relies only on RGB observations, language instructions, and interaction cues from the demonstrations, without using simulator object IDs or ground-truth target meshes.

\subsection{Instruction-Aware Spatial Representation Alignment}
\label{sec:representation_alignment}

After obtaining the instruction-aware targets, we need to transform their 3D spatial information into suitable supervision for VLA training. Since 3D meshes are variable-length, unordered, and lack a unified parameterization, we center and scale-normalize each reconstructed object mesh and render a fixed set of three orthographic views along its principal axes. The resulting canonical target views provide a consistent object-centric representation of the target geometry, reducing variations caused by camera viewpoint, scene pose, and visibility. As shown in Fig.~\ref{fig:architecture}, these views are then used for latent prediction and geometric feature alignment.

\paragraph{Canonical target-view latent prediction}
To reduce training overhead, we independently encode the three canonical views
using a frozen Stable Diffusion VAE~\cite{ldm} and stack the resulting latent
representations into the canonical target-view latent $\mathbf Z_m$. 
At each timestep, we additionally introduce a group of object-level queries. An
attention mask restricts their information interaction, allowing them to access
the language, proprioceptive-state, and visual tokens from the current and
historical timesteps. These object-level queries are then jointly decoded to predict the
latent representation $\hat{\mathbf Z}_t$ of the three canonical views. The
training objective is defined as
\begin{equation}
\mathcal L_{\mathrm{tri}}
=
\frac{1}{K}
\sum_{t=1}^{K}
\operatorname{MSE}\!\left(
\hat{\mathbf Z}_t,
\mathbf Z_{m_t}
\right),
\label{eq:tri}
\end{equation}
where $K$ denotes the number of observation timesteps,
$m_t\in\{1,\ldots,M\}$ is the active target index associated with
timestep $t$, and $\mathbf Z_{m_t}$ denotes the canonical target-view
latent of the corresponding target object. For single-target tasks,
$m_t=1$ for all timesteps. 

\paragraph{Target-object VGGT feature alignment}
Using solely canonical target-view latent prediction, Mind-VLA could miss explicit geometric supervision for intermediate VLA representations. Target-object VGGT feature alignment is thereby introduced to provide direct supervision of target-object 3D geometry and cross-view information. 
Specifically, the three canonical views are jointly fed into a frozen
VGGT~\cite{vggt}. At each selected representation level $j$, the patch
tokens of each view are mean-pooled and then averaged across the three
views to obtain the target-object feature $\mathbf{g}_{m,j}$. At the
corresponding VLA layer $j$, all visual tokens at timestep $t$ are
mean-pooled into an intermediate representation
$\bar{\mathbf{h}}_{t,j}$, which is mapped into the VGGT feature space
using a learnable projection $\phi_j$.
The geometric feature alignment
objective is defined as
\begin{equation}
\mathcal L_{\mathrm{geo}}
=
\frac{1}{JK}
\sum_{t=1}^{K}
\sum_{j=1}^{J}
\left[
1-
\cos\!\left(
\phi_j(\bar{\mathbf h}_{t,j}),
\mathbf g_{m_t,j}
\right)
\right],
\label{eq:geo}
\end{equation}
where $J$ denotes the number of aligned representation levels. 
Unlike Eq.~\eqref{eq:tri}, which supervises the object-level query output,
Eq.~\eqref{eq:geo} directly constrains intermediate VLA representations to
encode the 3D geometric structure and cross-view information associated with
the target object specified by the language instruction.

\paragraph{Temporal target assignment}
For tasks involving multiple sequential manipulation targets, we further assign
each timestep to its corresponding active target to construct the correct
target supervision. For grasping operations, target transitions are localized
according to the opening and closing cycles of the gripper; for pushing
operations, interaction stages are identified from significant motion changes
in CoTracker~\cite{cotracker} point trajectories. For tasks with more complex interaction
patterns, we jointly consider gripper states and object motion cues.
We then temporally match the detected interaction stages with the ordered
target sequence from Sec.~\ref{sec:target_acquisition} to determine
$m_t$. Once the current manipulation ends, supervision switches to the
next target from the following timestep, associating the approach motion with
the upcoming target.
\subsection{Overall Training and Inference}
\label{sec:overall}
We instantiate Mind-VLA on DreamVLA~\cite{dreamvla} as the base VLA model.
Following DreamVLA, we use the future-frame prediction objective
$\mathcal{L}_{\mathrm{pred}}$ and dynamic-region prediction objective
$\mathcal{L}_{\mathrm{dyn}}$ together with the action prediction objective
$\mathcal{L}_{\mathrm{act}}$.
The resulting base objective is defined as
\begin{equation}
\mathcal{L}_{\mathrm{base}}
=
\mathcal{L}_{\mathrm{act}}
+
\lambda_{\mathrm{pred}}\mathcal{L}_{\mathrm{pred}}
+
\lambda_{\mathrm{dyn}}\mathcal{L}_{\mathrm{dyn}}.
\end{equation}
We further introduce the proposed instruction-aware spatial objectives, and
the overall training objective of Mind-VLA is defined as
\begin{equation}
\mathcal{L}_{\mathrm{MindVLA}}
=
\mathcal{L}_{\mathrm{base}}
+
\lambda_{\mathrm{tri}}\mathcal{L}_{\mathrm{tri}}
+
\lambda_{\mathrm{geo}}\mathcal{L}_{\mathrm{geo}},
\end{equation}
where $\lambda_{\mathrm{tri}}$ and $\lambda_{\mathrm{geo}}$ control the
weights of canonical target-view latent prediction and target-object VGGT
feature alignment, respectively.

During training, Mind-VLA uses Instruction-Aware Target Acquisition and
Instruction-Aware Spatial Representation Alignment to provide
instruction-aware 3D supervision. At inference, the learned object-level
queries are retained, while instruction parsing, 3D reconstruction, VAE encoding,
and VGGT feature extraction are no longer required. Mind-VLA therefore
requires no additional 3D inputs or 3D preprocessing at inference time.

\section{EXPERIMENTS}

We evaluate Mind-VLA on LIBERO and CALVIN (Sec.~\ref{sec:exp_main}), analyze
the effects of instruction-aware spatial supervision and the learned
representations (Secs.~\ref{sec:alignment_analysis}
and~\ref{sec:representation_analysis}), and further evaluate robustness to
target occlusion in real-robot manipulation (Sec.~\ref{sec:real_robot}).

\subsection{Experimental Setup}

\paragraph{Base VLA and implementation}
We follow the model architecture and training settings of
DreamVLA~\cite{dreamvla} unless otherwise specified. Mind-VLA additionally
introduces a separate
group of object-level queries, forming an independent query branch from the
original Dream queries. During training, the original Dream queries and object-level queries are mutually masked, while the action queries can attend to both query branches, allowing the target-specific representations learned by the object-level queries to directly contribute to action prediction.

Mind-VLA uses nine object-level queries decoded by a two-layer MLP. For
geometric alignment, we use $J=4$ representation levels, pairing VLA layers
$\{3,6,9,11\}$ with VGGT layers $\{4,11,17,23\}$ through two-layer projection
heads (all 0-indexed). We set $\lambda_{\mathrm{tri}}=0.01$ and
$\lambda_{\mathrm{geo}}=0.1$. A single set of interaction thresholds is used
for all LIBERO-Long demonstrations. On an RTX 5090, the retained object-level
queries increase end-to-end action-chunk latency from 70.5 to 73.4\,ms.

\paragraph{Benchmarks and protocols}
We evaluate Mind-VLA on LIBERO~\cite{libero} and CALVIN ABC-D~\cite{calvin}, following the corresponding experimental protocols of
DreamVLA~\cite{dreamvla} on both benchmarks. LIBERO comprises Spatial, Object,
Goal, and Long suites, each with ten tasks and 50 demonstrations per task.
Models are pretrained on LIBERO-90 and finetuned separately for each suite,
and we report 20 rollouts per task. For CALVIN, we use the
video-prediction-pretrained initialization and language-annotated
demonstrations from environments A--C, and evaluate on the standard 1000
five-instruction sequences in the unseen environment D. We report success over
one to five consecutive tasks and the average completed length. The DreamVLA
results in Table~\ref{tab:calvin} are taken from the original
paper~\cite{dreamvla}.

\subsection{Main Benchmark Results}
\label{sec:exp_main}

Table~\ref{tab:libero} compares Mind-VLA with representative compact and large
VLA models on LIBERO. Using the same GPT-2-M backbone as DreamVLA, Mind-VLA
achieves 94.4\% average success, improving over the 92.6\% reported by
DreamVLA~\cite{dreamvla}. The largest gain appears on LIBERO-Object, where
success increases from 94.0\% to 98.0\%. With a compact 345M-parameter
backbone, Mind-VLA remains competitive with substantially larger VLA models.

\begin{table}[tb]
\vspace{6pt}
\centering
\caption{
\textbf{LIBERO benchmark results.} Success rate~(\%) on four
task suites. $^\dagger$Large VLAs (3--7\,B parameters) are
shown for reference; Mind-VLA uses a compact backbone
($\sim$345M). Best compact-backbone results are \textbf{bolded};
best overall are \underline{underlined}.
}
\label{tab:libero}
\setlength{\tabcolsep}{2pt}
\footnotesize
\begin{tabular}{l c c c c c c}
\toprule
\textbf{Method} & \textbf{Backbone} & \textbf{Spatial} & \textbf{Object} & \textbf{Goal} & \textbf{Long} & \textbf{Avg.} \\
\midrule
\multicolumn{7}{l}{\textit{Compact-backbone methods}} \\
Diff.\ Policy~\cite{diffusionpolicy}  & -- & 78.3 & 92.5 & 68.3 & 50.5 & 72.4 \\
MDT~\cite{mdt}                        & $<$300M & 78.5 & 87.5 & 73.5 & 64.8 & 76.1 \\
Octo~\cite{octo}                      & 93M & 78.9 & 85.7 & 84.6 & 51.1 & 75.1 \\
SmolVLA~\cite{smolvla}                & 0.45B & 90.0 & 96.0 & \textbf{92.0} & 71.0 & 87.3 \\
DreamVLA~\cite{dreamvla}              & 345M & 97.5 & 94.0 & 89.5 & \textbf{\underline{89.5}} & 92.6 \\
\textbf{Mind-VLA (Ours)}              & 345M & \underline{\textbf{98.0}} & \textbf{98.0} & \textbf{92.0} & \underline{\textbf{89.5}} & \underline{\textbf{94.4}} \\
\midrule
\multicolumn{7}{l}{\textit{Large VLAs$^\dagger$ (3--7\,B parameters)}} \\
OpenVLA~\cite{openvla}                & Llama-2-7B & 84.7 & 88.4 & 79.2 & 53.7 & 76.5 \\
TraceVLA~\cite{tracevla}              & Llama-2-7B & 84.6 & 85.2 & 75.1 & 54.1 & 74.8 \\
SpatialVLA~\cite{spatialvla}          & PaliGemma-4B & 88.2 & 89.9 & 78.6 & 55.5 & 78.1 \\
CoT-VLA~\cite{cotvla}                 & Llama-2-7B & 87.5 & 91.6 & 87.6 & 69.0 & 83.9 \\
CogACT~\cite{cogact}                  & Llama-2-7B & 97.2 & 98.0 & 90.2 & 88.8 & 93.6 \\
GLaD~\cite{glad}                      & Llama-2-7B & 95.0 & 97.4 & 94.4 & 89.4 & 94.1 \\
$\pi_0$~\cite{pi0}                   & PaliGemma-3B & 96.8 & \underline{98.8} & \underline{95.8} & 85.2 & 94.2 \\
\bottomrule
\end{tabular}
\end{table}

Table~\ref{tab:calvin} reports CALVIN ABC-D results. Mind-VLA reaches 79.4\%
success on Task 5 and an average completed length of 4.47, compared with
78.1\% and 4.44 reported by DreamVLA~\cite{dreamvla}, respectively.

\begin{table}[tb]
\centering
\caption{
\textbf{CALVIN ABC-D results.} Success rate for completing
1--5 consecutive tasks and average completed length
(Avg.\ Len.) over 1000 rollouts. DreamVLA results are taken from the original
paper~\cite{dreamvla}. $^\dagger$Large VLAs are shown for reference.
}
\label{tab:calvin}
\setlength{\tabcolsep}{2pt}
\footnotesize
\begin{tabular}{l c c c c c c}
\toprule
\textbf{Method} & \textbf{Task 1} & \textbf{Task 2} & \textbf{Task 3} & \textbf{Task 4} & \textbf{Task 5} & \textbf{Avg.\ Len.} \\
\midrule
GR-1~\cite{gr1}                   & 85.4 & 71.2 & 59.6 & 49.7 & 40.1 & 3.06 \\
OpenVLA$^\dagger$~\cite{openvla}  & 91.3 & 77.8 & 62.0 & 52.1 & 43.5 & 3.27 \\
CLOVER~\cite{clover}              & 96.0 & 83.5 & 70.8 & 57.5 & 45.4 & 3.53 \\
$\pi_0^\dagger$~\cite{pi0}       & 93.8 & 85.0 & 76.7 & 68.1 & 59.9 & 3.84 \\
Seer~\cite{seer}                  & 96.3 & 91.6 & 86.1 & 80.3 & 74.0 & 4.28 \\
VPP~\cite{vpp}                    & 95.7 & 91.2 & 86.3 & 81.0 & 75.0 & 4.29 \\
DreamVLA~\cite{dreamvla}          & \textbf{98.2} & 94.6 & 89.5 & 83.4 & 78.1 & 4.44 \\
\textbf{Mind-VLA (Ours)}          & 98.0 & \textbf{95.2} & \textbf{89.9} & \textbf{84.6} & \textbf{79.4} & \textbf{4.47} \\
\bottomrule
\end{tabular}
\end{table}

\subsection{Instruction-Aware Spatial Alignment Analysis}
\label{sec:alignment_analysis}

Table~\ref{tab:ablation} provides controlled comparisons of scene-level and
target-object spatial supervision under the same backbone, pretrained checkpoint, base training objective, and optimization settings.

\begin{table*}[tb]
\vspace*{6pt}
\centering
\caption{\textbf{Controlled LIBERO ablation.} All variants use the same backbone, pretrained checkpoint, base training objective, and optimization recipe. A4 and A5 differ only in the VGGT alignment target, and A6 adds temporal
target assignment to A5.}
\label{tab:ablation}
\setlength{\tabcolsep}{4.0pt}
\footnotesize
\begin{tabular}{c l c c c c c c c}
\toprule
ID & Configuration & $\mathcal L_{\rm tri}$ & VGGT target & Spatial & Object & Goal & Long & Avg.\\
\midrule
A0 & Base model & -- & --
&97.0&93.5&89.5&85.0&91.3\\
A1 & $\mathcal L_{\rm tri}$ only & \checkmark & --
&97.5&95.0&91.0&86.5&92.5\\
A2 & Matched SF reimplementation & -- & Scene RGB
&\textbf{98.0}&92.5&90.0&89.0&92.4\\
A3 & $\mathcal L_{\rm geo}$ only & -- & Canonical target views
&\textbf{98.0}&97.5&91.5&87.0&93.5\\
A4 & $\mathcal L_{\rm tri}$ + scene-VGGT control & \checkmark & Scene RGB
&97.5&94.0&89.5&\textbf{90.0}&92.8\\
A5 & Mind-VLA w/o target assignment & \checkmark & Canonical target views
&\textbf{98.0}&\textbf{98.0}&\textbf{92.0}&87.5&93.9\\
\rowcolor{gray!18}
A6 & Mind-VLA full & \checkmark & Canonical target views
&\textbf{98.0}&\textbf{98.0}&\textbf{92.0}&89.5&\textbf{94.4}\\
\bottomrule
\end{tabular}
\end{table*}

\paragraph{Target-object alignment}
We first isolate the effect of replacing scene-level geometric supervision
with instruction-aware target-object supervision. Without
$\mathcal L_{\rm tri}$, replacing scene-level VGGT features with canonical target-view
features improves the average success from 92.4\% to 93.5\%
(A2$\rightarrow$A3), including a gain from 92.5\% to 97.5\% on LIBERO-Object.
The same trend holds with canonical target-view latent prediction: A4 and A5 differ only in
the VGGT alignment target, and target-object alignment improves the average
from 92.8\% to 93.9\%. These comparisons show that the improvement comes from
making spatial supervision instruction-aware rather than merely introducing
VGGT alignment.
\paragraph{Target selection and canonical geometry}
We then separate the effects of instruction-aware target selection and canonical target geometry.
On LIBERO-Object, scene-VGGT alignment achieves 94.0\% success.
In an additional target-crop control, replacing scene-VGGT features with features from the instruction-specified target object improves the success rate to 96.5\%, showing the benefit of focusing spatial supervision on the instructed target. Using canonical target-view alignment further improves the success rate to 98.0\%, suggesting that canonical target views provide more consistent target-object spatial supervision by reducing variations caused by camera viewpoint, scene pose, and visibility.

\paragraph{Supervision objectives and temporal assignment}
We further examine how the proposed supervision objectives and temporal assignment contribute to performance. The two target-object objectives provide complementary supervision, with $\mathcal{L}_{\mathrm{tri}}$ supervising object-level query predictions and $\mathcal{L}_{\mathrm{geo}}$ constraining intermediate VLA representations. Their combination achieves 93.9\% average success (A5). Temporal target assignment further improves LIBERO-Long from 87.5\% to 89.5\% (A5$\rightarrow$A6), supporting its role in sequential multi-target manipulation.

\paragraph{Automatic supervision construction}
 We additionally evaluate the automatically constructed supervision on LIBERO from instruction parsing, target selection, and temporal consistency. The target sequences are correctly parsed and the automatic pipeline correctly selects the manipulated target for all 40 tasks. On LIBERO-Long, 494 of 500 demonstrations pass the phase-consistency check between the detected interaction events and parsed target phases. These results show that the automatic pipeline can reliably identify instruction-specified targets and construct temporally consistent supervision.
\begin{table}[tb]
\centering
\caption{\textbf{Real-robot success rate (\%).} Thirty trials per task. The
scene-VGGT variant is the matched control, sharing Mind-VLA's architecture and
losses and differing only in the VGGT alignment target; OpenVLA and Seer are
external references.}
\label{tab:real_robot}
\setlength{\tabcolsep}{2.7pt}
\small
\begin{tabular}{l c c c c}
\toprule
\textbf{Task} & \textbf{OpenVLA} & \textbf{Seer} & \textbf{Scene-VGGT} & \textbf{Mind-VLA}\\
\midrule
\multicolumn{5}{l}{\textit{Normal}}\\
Pick--Banana & 33 & 43 & 63 & \textbf{70}\\
Pick--Potato & 10 & 30 & 37 & \textbf{57}\\
Drawer & 40 & 63 & 70 & \textbf{73}\\
Place--Banana & 27 & 67 & 70 & \textbf{80}\\
Place--Potato & 20 & 43 & 47 & \textbf{63}\\
\midrule
\multicolumn{5}{l}{\textit{Occluded ($\sim$25\%)}}\\
Pick--Banana & 3 & 23 & 27 & \textbf{57}\\
Pick--Potato & 7 & 13 & 20 & \textbf{43}\\
Drawer & 10 & 30 & 37 & \textbf{63}\\
\bottomrule
\end{tabular}
\end{table}
\subsection{Representation Analysis}
\label{sec:representation_analysis}
To examine whether Mind-VLA learns target-specific geometric representations, we freeze each model and train layer-wise linear probes under a ten-fold leave-one-object-out protocol.
Target geometry decodability is measured by centered cosine similarity to the held-out target descriptor, while target association is evaluated using ten-way top-1 retrieval accuracy.
DreamVLA, Mind-VLA without $\mathcal{L}_{\mathrm{geo}}$, and a Vanilla VLA are included as controls.

\begin{figure*}[tb]
\vspace*{6pt}
\centering
\includegraphics[width=0.90\textwidth]{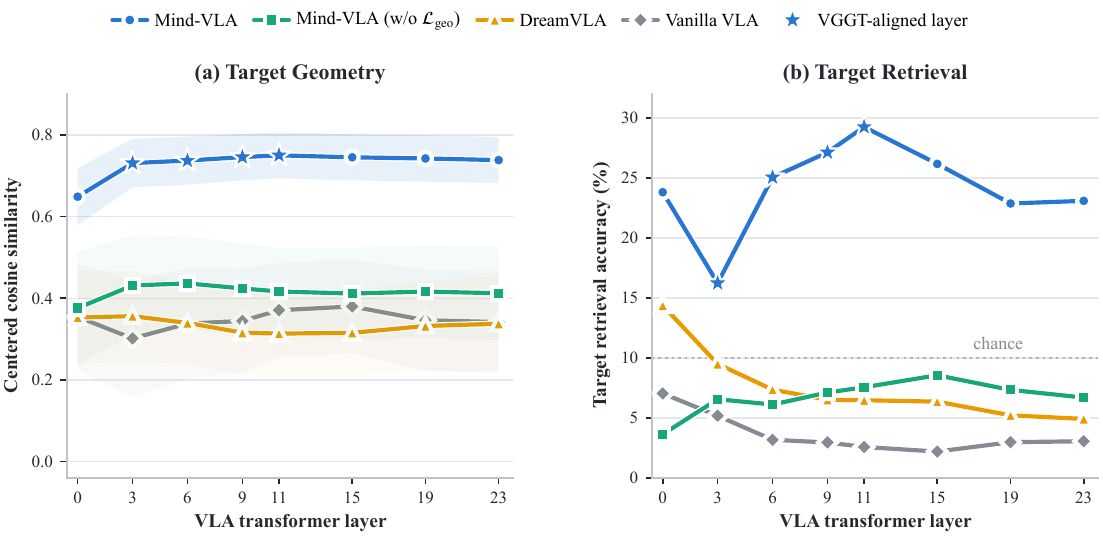}
\caption{\textbf{Layer-wise representation analysis of Mind-VLA.}
Layer-wise linear probes are trained under a ten-fold leave-one-object-out
protocol. (a) Target geometry is evaluated by centered cosine similarity to
the held-out target descriptor; shaded bands show $\pm1$ s.e.m.
(b) Target retrieval is evaluated by ten-way top-1 accuracy, with 10\% chance
accuracy. Star markers indicate the Mind-VLA layers directly aligned with VGGT
features by $\mathcal L_{\rm geo}$.}
\label{fig:representation_probe}
\end{figure*}
\paragraph{Target geometry}
As shown in Fig.~\ref{fig:representation_probe}(a), Mind-VLA encodes substantially more target-geometry information than the baselines, reaching a centered cosine similarity of 0.75 at layer 11, while DreamVLA and Vanilla VLA remain around 0.3--0.4. Removing $L_{\mathrm{geo}}$ markedly reduces the decodability, validating the effect of target-object geometric alignment.

\paragraph{Target retrieval}
We then examine whether the decoded geometry is associated with the instruction-specified target. As shown in Fig.~\ref{fig:representation_probe}(b), Mind-VLA achieves 29\% top-1 target retrieval accuracy, substantially above the 10\% chance level. Removing the target identity from the instruction reduces geometry decodability to 0.41--0.45 and retrieval accuracy to 2--4\%, demonstrating that the learned geometric representation is instruction-aware.
\begin{figure*}[!t]
\vspace*{6pt}
\centering
\begin{minipage}[c]{0.7\textwidth}
    \centering
    \includegraphics[width=\linewidth]{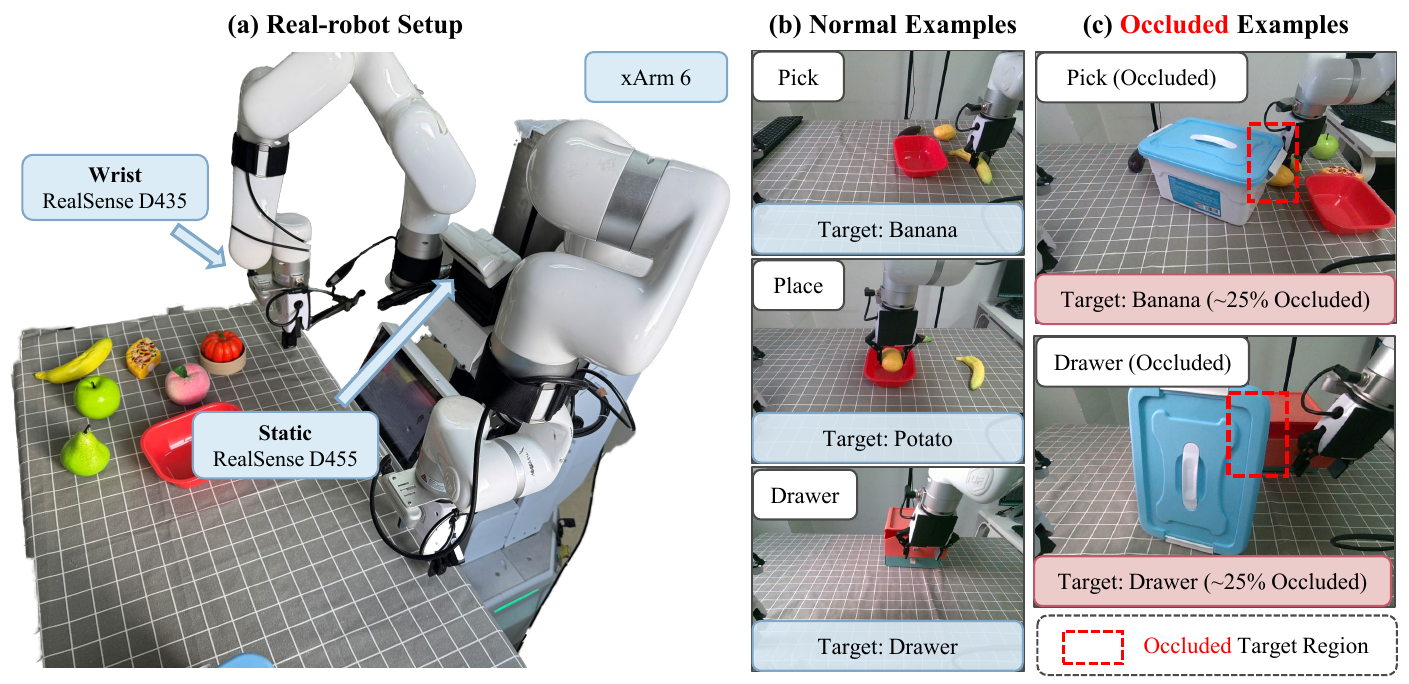}
    \vspace{-1mm}
\end{minipage}\hfill
\begin{minipage}[c]{0.295\textwidth}
    \centering
    \includegraphics[width=\linewidth]{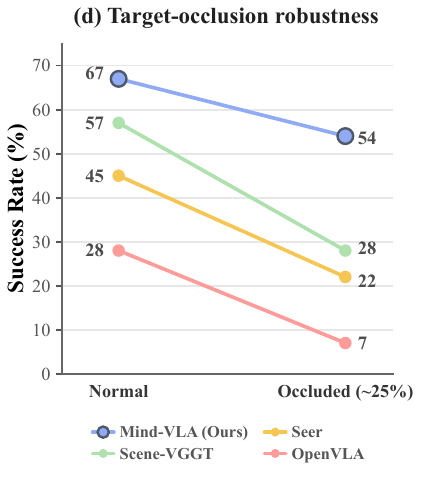}
    \vspace{-1mm}
\end{minipage}
\caption{\textbf{Real-robot evaluation.}
(a) Robot setup. (b) Normal task examples.
(c) Examples under approximately 25\% target occlusion.
(d) Robustness to target occlusion.
The averages in (d) are computed over the three tasks evaluated under both normal and occluded settings.}
\label{fig:real_robot}
\end{figure*}

\subsection{Real-Robot Occlusion Robustness}
\label{sec:real_robot}
Finally, we examine whether the learned instruction-aware geometry translates into robust manipulation under real-world target occlusion (Fig.~\ref{fig:real_robot}). We use one arm of a dual xArm 6 setup with a static RealSense D455 and a wrist-mounted D435 (Fig.~\ref{fig:real_robot}(a)). Following DreamVLA~\cite{dreamvla}, we first pretrain the base VLA on DROID~\cite{droid} without auxiliary objectives. Mind-VLA and the matched scene-VGGT control are initialized from this pretrained model. OpenVLA and Seer are initialized from their respective released pretrained checkpoints. All methods are then finetuned on the same 50 demonstrations per task following their respective training recipes for fair comparison. 

We evaluate Pick, Place, and Drawer, with 30 trials per
reported cell. Before each trial, the robot is reset to the same preset
initial arm pose, while the task objects are randomly repositioned within
predefined task-specific workspace regions. Pick succeeds when the specified object is held
stably for at least 10\,s; Place when it reaches the designated region; and
Drawer when it is fully closed. A wrong-object grasp counts as failure.
Occluded Pick and Drawer use a physical occluder that blocks approximately
25\% of the target, and any arm--occluder contact counts as failure.

As shown in Table~\ref{tab:real_robot} and Fig.~\ref{fig:real_robot}(d), Mind-VLA achieves 54\% average success under approximately 25\% target occlusion, outperforming the matched scene-VGGT control by 26 percentage points (54\% vs. 28\%). From normal to occluded settings, Mind-VLA drops by only 13 points (67\% to 54\%), while the scene-VGGT control drops by 29 points (57\% to 28\%). Seer and OpenVLA also show larger degradation, decreasing from 45\% to 22\% and from 28\% to 7\%, respectively. These results demonstrate the advantage of instruction-aware target geometry for robust manipulation under target occlusion.

\section{CONCLUSION}

To address manipulation failures caused by instruction-agnostic 3D modeling, we propose Mind-VLA, which aligns VLA representations with the 3D geometry of instruction-specified targets rather than the entire scene. Mind-VLA uses canonical target geometry to provide instruction-aware spatial supervision during training, without requiring additional 3D inputs or 3D preprocessing at inference. Experiments on LIBERO, CALVIN, and real robots demonstrate improved manipulation performance and stronger robustness to target occlusion. These results highlight the importance of instruction-aware spatial supervision for  VLA learning.

\bibliographystyle{IEEEtran}
\bibliography{main}

\end{document}